\documentclass{article}

\usepackage[preprint]{neurips_2026}
\usepackage{natbib}
\setcitestyle{authoryear}

\usepackage[utf8]{inputenc}
\usepackage[T1]{fontenc}
\usepackage{amsmath, amssymb, amsthm}
\usepackage{graphicx}
\usepackage{booktabs}
\usepackage{algorithm}
\usepackage{algorithmic}
\usepackage{hyperref}
\usepackage{url}
\usepackage{xcolor}
\usepackage{nicefrac}
\usepackage{microtype}
\usepackage{subcaption}
\usepackage{multirow}
\usepackage{wrapfig}

\definecolor{bluecite}{HTML}{0875b7}
\hypersetup{
  colorlinks=true,
  citecolor=bluecite,
  linkcolor=magenta
}

\title{Self-Supervised On-Policy Reinforcement Learning via Contrastive Proximal Policy Optimisation}

\author{
  Asim Osman\thanks{Equal contribution. Correspondence: \texttt{a.osman@instadeep.com}} \\ InstaDeep, AIMS
  \And Sasha Abramowitz\footnotemark[1] \\ InstaDeep
  \And Mark Bergh \\ InstaDeep
  \And Ulrich Armel Mbou Sob \\ InstaDeep
  \And Ruan John de Kock \\ InstaDeep
  \And Omayma Mahjoub \\ InstaDeep
  \And Oussama Hidaoui \\ InstaDeep
  \And Noah De Nicola \\ InstaDeep
  \And Arnol Manuel Fokam \\ InstaDeep
  \And Felix Chalumeau \\ InstaDeep
  \And Daniel Rajaonarivonivelomanantsoa \\ InstaDeep, University of Stellenbosch
  \And Siddarth Singh \\ InstaDeep
  \And Refiloe Shabe \\ InstaDeep
  \And Juan Claude Formanek \\ InstaDeep
  \And Simon Verster Du Toit \\ InstaDeep
  \And Arnu Pretorius \\ InstaDeep
}

\begin{document}
\maketitle
\begin{abstract}

Contrastive reinforcement learning (CRL) learns goal-conditioned Q-values through a contrastive objective over state-action and goal representations, removing the need for hand-crafted reward functions. Despite impressive success in achieving viable self-supervised learning in RL, all existing CRL algorithms rely on off-policy optimisation and are mostly constrained to continuous action spaces, with little research invested in discrete environments. This leaves CRL disconnected from widely used and effective, modern on-policy training pipelines adopted across both single-agent and multi-agent RL in continuous and discrete environments. To establish a first connection, we introduce Contrastive Proximal Policy Optimisation (CPPO). CPPO is an on-policy contrastive RL algorithm that derives policy advantages directly from contrastive Q-values and optimises them via the standard PPO objective, without requiring a reward function or a replay buffer. We evaluate CPPO across continuous and discrete, single-agent and cooperative multi-agent tasks.  Whilst the existence of an on-policy approach is inherently useful, we observe that \textbf{CPPO not only significantly outperforms the previous CRL baselines in 14 out of 18 tasks, but also matches or exceeds PPO's performance, which uses hand-crafted dense rewards, in 12 out of the 18 tasks tested.}

\end{abstract}


\section{Introduction}

Designing reward functions is one of the most fragile parts of applying reinforcement learning in practice. Sparse rewards make exploration intractable, dense rewards introduce specification bias and reward hacking, and small misspecifications can produce degenerate behaviour \citep{reward-shaping-challenges,reward-fragility,reward-hacking}. Contrastive reinforcement learning (CRL)~\citep{eysenbach2022,single-goal-crl}, sidesteps this entirely: rather than maximising a scalar reward, the agent learns to reach goal states, using a $Q$-function trained via a contrastive objective. Since its introduction, CRL has matured rapidly, with progress in emergent exploration, scalable training, deeper architectures, and multi-agent extensions \citep{jaxgcrl,icrl,1000layer}


Yet every existing CRL method to date shares an underlying design choice: they are all off-policy, built on SAC~\citep{sac} or DQN~\citep{dqn}. This reflects a design choice inherited from CRL's SAC-based origins that has propagated through the literature, leaving the contrastive paradigm without an on-policy counterpart. In reward-based RL, on-policy and off-policy methods have developed as parallel families with complementary strengths: DQN~\citep{dqn} alongside A3C~\citep{a2c}, SAC~\citep{sac} alongside PPO~\citep{ppo}, MADDPG~\citep{maddpg} alongside IPPO~\citep{ippo}, in the multi-agent setting. Off-policy methods offer sample efficiency through replay; on-policy methods offer stability, simpler implementation ~\citep{sutton-barto}, and strong compatibility with massively parallel simulation~\citep{dpo,jumanji,mava}, where PPO has become the standard choice~\citep{isaacgym, rudin2021minutes}. In cooperative MARL on-policy methods have become the prevalent paradigm~\citep{MAT, mappo, sable}. The contrastive RL literature has no counterpart for on-policy learning, meaning practitioners who want a goal-conditioned method must accept the instabilities and design constraints of off-policy learning, even in settings where on-policy methods would otherwise be the natural choice.


In this work, we introduce Contrastive Proximal Policy Optimisation (CPPO)\footnote{See our \href{https://sites.google.com/view/contrastive-ppo/home}{website} for the implementation and hyperparameters}, the first on-policy contrastive RL algorithm. CPPO derives policy advantages directly from a contrastive $Q$-function and optimises its policy via PPO's clipped surrogate objective. CPPO is a self-supervised method that handles discrete action spaces natively, extends cleanly to the MARL setting, and integrates into existing on-policy training pipelines with minimal modification. Of the 18 tasks benchmarked, CPPO significantly outperforms contrastive baselines on 14 and, despite using no reward signal, it matches or exceeds the performance of reward-based PPO on 12.

\textbf{Our contributions are as follows:}
\begin{enumerate}
\item \textbf{CPPO}, an on-policy contrastive RL algorithm that computes advantages directly from contrastive Q-values and optimises them via PPO's clipped objective, requiring no reward function, replay buffer or target network.
\item \textbf{A general method} that operates across single-agent and multi-agent, discrete and continuous settings without significant algorithmic modifications, while outperforming CRL baselines in discrete settings.

\item \textbf{Evaluation of contrastive RL against dense-reward baselines.} Prior CRL work compares primarily against sparse-reward or goal-conditioned methods; we benchmark against PPO with dense-rewards, and contrast goal engineering with reward engineering.
\end{enumerate}


\begin{table}[t]
\centering
\caption{\textbf{\textit{Design comparison of contrastive RL algorithms}}.
We highlight with $^*$ the method naming convention used in our experiments. It aims to remove ambiguity by connecting the underlying base RL algorithm to the contrastive learning approach. 
In our work, CPPO, is a self-supervised RL method with a PPO base that is on-policy and naturally suited for both discrete and continuous action spaces. We categorise prior work similarly.}
\label{tab:method-comparison}
\resizebox{0.9\columnwidth}{!}{%
\begin{tabular}{llcccc}
\toprule
\textbf{Source} & \textbf{Method}* & \textbf{On-Policy} & \textbf{Discrete} & \textbf{Continuous} & \textbf{Multi-Agent}\\
\midrule
CRL~\citep{eysenbach2022}              & CSAC & \texttimes & \texttimes & \checkmark & \texttimes\\
CRL~\citep{demystifying-sgcrl} & CDQN & \texttimes  & \checkmark & \texttimes & \texttimes \\
ICRL~\citep{icrl}                      & ICSAC & \texttimes & \checkmark\rlap{\textsuperscript{$\dagger$}} & \checkmark & \checkmark \\
\midrule
\textbf{Ours}                  & I/CPPO & \checkmark & \checkmark & \checkmark & \checkmark \\
\bottomrule
\multicolumn{6}{l}{\small $^\dagger$Via Gumbel-Softmax approximation.} \\
\end{tabular}%
}
\end{table}

\section{Background}

\paragraph{Problem Formulation}

We model the single-agent reinforcement learning (RL) problem as a partially observable Markov decision process (POMDP), defined by the tuple
$\langle \mathcal{S}, \mathcal{A}, \mathcal{O}, P, R, \gamma \rangle$, where
$\mathcal{S}$, $\mathcal{A}$, and $\mathcal{O}$ denote the state, action, and
observation spaces respectively. At each timestep, the agent receives an
observation $o \in \mathcal{O}$, selects an action $a \in \mathcal{A}$, and the
environment transitions to a new state according to the transition function
$P: \mathcal{S} \times \mathcal{A} \rightarrow \Delta(\mathcal{S})$, producing
a scalar reward $r = R(s, a)$ and a new observation. The agent's objective is
to learn a policy $\pi(a | o)$ that maximises the expected discounted return
$J(\pi) = \mathbb{E}\left[\sum_{t=0}^{T} \gamma^t r_t\right]$ where $T \in \mathbb{N}$ is the episode horizon.

We model the cooperative multi-agent reinforcement learning (MARL) problem as a decentralised-POMDP (Dec-POMDP) ~\citep{dec-pomdp} where $n$ agents act simultaneously. At each step, agent
$i$ selects an action $a_i$ based on its local observation $o_i$, forming a
joint action $\mathbf{a} = (a_1, \dots, a_n)$ that transitions the environment
according to $P$. In cooperative settings, all agents share a common reward
$R: \mathcal{S} \times \mathcal{A}^n \rightarrow \mathbb{R}$. A popular approach is independent learning, where each agent optimises its own policy
$\pi_i(a_i | o_i)$ using the shared reward signal~\citep{ippo}.

\paragraph{Goal-Conditioned RL}
\label{sec:gcrl}
Goal-conditioned reinforcement learning (GCRL) replaces the scalar reward with a goal: the agent receives a target state $g \in \mathcal{S}$ and must learn a policy $\pi(a \mid s, g)$ that reaches the goal~\citep{Kaelbling,uvf-schaul15}. Rather than maximising cumulative reward, the objective is to maximise the $\gamma$-discounted state occupancy measure at the goal~\citep{eysenbach2022,single-goal-crl}:
\begin{equation}
    \max_\pi \; \rho^\pi_\gamma(g), \qquad
    \rho^\pi_\gamma(g) \triangleq (1 - \gamma) \sum_{t=0}^{\infty}
    \gamma^t \, p^\pi_t(s_t = g),
    \label{eq:gcrl-objective}
\end{equation}
where $p^\pi_t(s_t = g)$ is the probability of visiting state $g$ at
time $t$ under policy $\pi$. The corresponding goal-conditioned
Q-function is the conditional occupancy measure:
\begin{equation}
    Q^\pi(s, a, g) \triangleq \rho^\pi_\gamma(g \mid s, a).
    \label{eq:gcrl-q}
\end{equation}
In practice, we follow ~\cite{single-goal-crl} and condition on a single fixed target goal $g^*$ rather than sampling from a goal distribution, which has been shown to be sufficient to learn goal-conditioned policies via contrastive RL~\citep{single-goal-crl,demystifying-sgcrl,icrl}.

\paragraph{Contrastive RL}
\label{sec:crl}

The goal-conditioned Q-function (Equation~\ref{eq:gcrl-q}) forms the basis of CRL. It is parameterised by two encoders: a state-action encoder $\phi(s, a)$ and a goal
encoder $\psi(g)$, both mapping to a $d$-dimensional representation
space. A critic is defined as a similarity function over these
representations and is trained to approximate the goal conditioned Q-function.



The encoders are trained using the InfoNCE objective~\citep{infonce} on
batches of trajectories. For each state-action pair $(s_i, a_i)$, a
positive goal $g_i^+$ is sampled from a future state the policy reached in the same trajectory, a procedure known as hindsight relabeling (HER) ~\citep{her}, while negative goals $g_j^-$ are drawn from other trajectories:
\begin{equation}
    \mathcal{L}_{\text{InfoNCE}} = -\frac{1}{N} \sum_{i=1}^{N} \log
    \frac{\exp(f(s_i, a_i, g_i^+))}
    {\sum_{j=1}^{N} \exp(f(s_i, a_i, g_j^-))}
    \label{eq:infonce}
\end{equation}
Thus the critic is trained to map state-action pairs and achieved goals to a shared latent space where their representations are aligned. A similarity measure over these representations serves as a proxy for the probability of reaching the goal. Our method uses negative L2 distance (see Equation~\ref{eq:goal-conditioned-q}) but other distance measures are common in the literature such as taking the inner product~\citep{jaxgcrl}. Thus, this is equivalent to the $Q$-function described by Equation~\ref{eq:gcrl-q}. 

Prior work takes inspiration from SAC~\citep{sac} when training the actor. Simply training the actor to pick the action which maximises it's future value.
%

\paragraph{Multi-Agent CRL}
\label{sec:ma-crl}

Independent Contrastive RL (ICRL)~\citep{icrl} extends CRL to cooperative multi-agent settings by reframing the Dec-POMDP as a joint goal-reaching problem. A mapping $m_g \colon \mathcal{O}^{1:n} \to \mathcal{G}$ produces a goal representation from joint observations, with a single fixed target $g^*$~\citep{single-goal-crl,icrl}. The team's objective is then to maximise the joint state-occupancy of $g^*$. 

ICRL follows the independent learning paradigm with parameter sharing: each agent acts on local observations $o_i$ but all agents share the policy $\pi_\theta$ and the contrastive encoders $\phi_\xi, \psi_\omega$. Like single-agent CRL, it builds on an off-policy SAC backbone, but uses Straight-Through Gumbel-Softmax~\citep{gumbel-softmax} to handle discrete actions through a continuous relaxation.


\paragraph{Proximal Policy Optimisation}
\label{sec:ppo}

PPO~\citep{ppo} is a policy gradient
algorithm that stabilises training by constraining each policy update to a
trust region. Given a batch of transitions collected under a behaviour
policy $\pi_{\theta_\text{old}}$, PPO optimises the clipped surrogate
objective
\begin{equation}
  L^\text{CLIP}(\theta) = \mathbb{E}_t \bigl[
    \min\!\bigl(
      r_t(\theta)\, \hat{A}_t,\;
      \text{clip}(r_t(\theta),\, 1{-}\epsilon,\, 1{+}\epsilon)\, \hat{A}_t
    \bigr)
  \bigr],
  \label{eq:ppo-clip}
\end{equation}
where $r_t(\theta) = \frac{\pi_\theta(a_t \mid o_t)}{\pi_{\theta_\text{old}}(a_t \mid o_t)}$
is the probability ratio and $\epsilon$ is a clipping threshold. The advantages $\hat{A}_t$ are estimated via Generalised Advantage
Estimation (GAE)~\citep{gae}, which requires a learned value function
$V_\psi(o_t)$ and per-step rewards:
\begin{equation}
  \hat{A}_t^\text{GAE} = \sum_{l=0}^{T-t} (\gamma \lambda)^l \,
    \delta_{t+l}, \qquad
  \delta_t = r_t + \gamma\, V_\psi(o_{t+1}) - V_\psi(o_t).
  \label{eq:gae}
\end{equation}
In the multi-agent setting, IPPO~\citep{ippo} applies PPO independently to
each agent with shared parameters, while MAPPO~\citep{mappo} additionally
conditions the value function on the global state. Both have become two of the
dominant on-policy optimisation approaches in cooperative
MARL~\citep{mappo, mava, jaxmarl}.
\section{Method}
\label{sec:method}

We introduce Contrastive Proximal Policy Optimisation (CPPO), our approach to on-policy contrastive reinforcement learning. We describe: (1) how we use contrastive learning to estimate the advantage, (2) how we tie this into PPO's existing optimisation objective, and (3) how we can extend CPPO to the multi-agent setting.

\subsection{Advantage Estimation Without Rewards}
\label{sec:adv-est-without-rewards}

CPPO is a policy gradient algorithm that uses contrastive Q-values to compute advantages, replacing the standard GAE~\citep{gae} that requires a learned value function and reward signal. Prior CRL methods train the actor by directly maximising the critic~\citep{eysenbach2022, icrl, 1000layer}: $\max_\pi \; \mathbb{E}_{a \sim \pi}\left[Q(s, a)\right]$ by following either SAC-style policies learned through critic hill-climbing~\citep{sac} or DQN style $\epsilon$-greedy policies~\citep{dqn}. Instead, we use the critic to compute advantages for PPO's clipped objective, with advantage defined as $A^\pi(s, a) = Q^\pi(s, a) - V^\pi(s)$, where $V^\pi(s) = \mathbb{E}_{a \sim \pi}[Q^\pi(s, a)]$.





The advantage formulation above only requires access to a $Q$ function. To instantiate it in the reward-free regime, we obtain $Q$ through contrastive learning, which replaces rewards with goal-reaching as the supervision signal. Concretely, CPPO uses three networks:

\begin{itemize}
    \item Policy network: $\pi_\theta(a \mid o, g)$ that maps observations
    and goals to action distributions.
    \item State-action encoder: $\phi_\xi(o, a)$ that maps observation-action
    pairs to a $d$-dimensional representation space.
    \item Goal encoder: $\psi_\omega(g)$ that maps goals to the same
    $d$-dimensional space.
\end{itemize}

The combination of the state-action encoder and the goal encoder form our $Q$ function as described in Section~\ref{sec:crl}. Therefore, in the goal conditioned regime, we can compute advantages as follows, always using the single fixed target goal $g^*$:
\begin{enumerate}
    \item Compute Q-values for the chosen action using the encoders:
    \begin{equation}
        Q(o, a, g^*) = -\|\phi(o, a) - \psi(g^*)\|_2 \quad 
        \label{eq:goal-conditioned-q}
    \end{equation}

\item Compute the value of the current state. For discrete action spaces, we compute the expectation over all actions:
\begin{equation}
    V(o, g^*) = \mathbb{E}_{a \sim \pi(\cdot \mid o,g^*)}[Q(o,a,g^*)] =  \sum_{a \in \mathcal{A}} \pi(a \mid o, g^*) \cdot Q(o, a, g^*).
    \label{eq:value-discrete}
\end{equation}
For continuous action spaces, the value is the integral over all actions \mbox{$V(o, g^*) = \int_{\mathcal{A}} \pi(a \mid o, g^*)Q(o, a, g^*) da$}. Since this is intractable in practice, we estimate it via Monte Carlo sampling with $K$ actions drawn from the policy:
\begin{equation}
    V(o, g^*) \approx \frac{1}{K} \sum_{k=1}^{K} Q(o, a_k, g^*), \quad a_k \sim \pi(\cdot \mid o, g^*).
    \label{eq:value-continuous}
\end{equation}

    \item Compute advantages:
    \begin{equation}
        A(o, a, g^*) = Q(o, a, g^*) - V(o, g^*)
        \label{eq:advantage}
    \end{equation}
\end{enumerate}

\subsection{Contrastive Proximal Policy Optimisation}
\label{sec:cppo}

Now that we have defined a method to estimate advantages using contrastive learning, we incorporate this directly into the PPO algorithm in a straightforward way. Each training iteration consists of two phases:
\begin{enumerate}
    \item \textbf{Encoder update}: update $\phi$ and $\psi$ using the InfoNCE loss (Equation~\ref{eq:infonce}) on hindsight-relabeled data. Goals are relabeled using future observations from the same episode, with negative samples drawn from other episodes in the batch.
    \item \textbf{Policy update}: compute contrastive advantages $A$ using the method defined in Section~\ref{sec:adv-est-without-rewards}, then update $\pi$ using the PPO clipped surrogate objective (Equation \ref{eq:ppo-clip}).
\end{enumerate}

The complete CPPO algorithm is summarised in Algorithm~\ref{alg:CPPO}.

    
\begin{algorithm}[t]
\caption{CPPO: On-Policy Contrastive RL}
\label{alg:CPPO}
\begin{algorithmic}[1]
\small
\STATE Initialise policy $\pi_\theta$, encoders $\phi_\xi$, $\psi_\omega$
\FOR{iteration $= 1, \dots, N$}
    \STATE Collect rollouts with $\pi_\theta$ across $M$ parallel environments
    \STATE Relabel (with HER): sample future observation $o_{t+k}$ from the same trajectory as goal $g$
    \STATE $Q(o,a,g^*) = f\!\bigl(\phi_\xi(o,a),\, \psi_\omega(g^*)\bigr)$,\quad $V(o,g^*) = \textstyle\sum_{a} \pi_\theta(a\mid o,g^*)\, Q(o,a,g^*)$ \unskip\hfill $\triangleright$ Eqs.~\ref{eq:goal-conditioned-q},\ref{eq:value-discrete}
    \STATE $A = Q - V$ \unskip\hfill $\triangleright$ Eq.~\ref{eq:advantage}
    \FOR{epoch $= 1, \dots, K$}
        \STATE Update $\phi_\xi, \psi_\omega$ via $\mathcal{L}_{\mathrm{InfoNCE}}$ \unskip\hfill $\triangleright$ Eq.~\ref{eq:infonce}
        \STATE Update $\pi_\theta$ via $\mathcal{L}_{\mathrm{PPO}}$ with advantages $A$ \unskip\hfill$\triangleright$ Eq.~\ref{eq:ppo-clip}
    \ENDFOR
\ENDFOR
\end{algorithmic}
\end{algorithm}

\subsection{Multi-Agent CPPO}
\label{sec:ma-cppo}

Cooperative MARL is a setting where reward shaping is particularly fragile: dense team rewards must implicitly solve the credit assignment problem across agents, and small misspecifications can produce degenerate cooperative behaviour. This makes it a natural target for reward-free methods. ICRL~\citep{icrl} was the first to demonstrate that contrastive RL extends to cooperative settings. It inherits the off-policy SAC backbone from single-agent CRL and relies on Gumbel-Softmax to handle discrete actions. We instead extend CPPO to the multi-agent case, yielding an on-policy contrastive MARL algorithm compatible with the IPPO/MAPPO training paradigm that is widely adopted in the field~\citep{ippo,happo, MAT, mappo, sable}.

We adopt an IPPO-style independent learning formulation with full parameter sharing: all agents share the policy $\pi_\theta$ and the encoders $\phi_\xi$, $\psi_\omega$, with agent identity supplied via the observation. Following ICRL, all agents condition on a single shared goal $g^*$, and the InfoNCE loss draws negatives from other trajectories in the batch. We refer to the independent version of multi-agent CPPO as ICPPO. A centralised critic analogue in the spirit of MAPPO is a natural extension to consider for future work.

\section{Experiments}
\label{sec:experiments} 
 
 \subsection{Experimental design}
 

\paragraph{Environments}
Our experiments include discrete and continuous, single- and multi-agent environments. Specifically, we evaluate CPPO/ICPPO in the following JAX-based environments: Navix~\citep{navix} \{single-agent, discrete\}, JaxGCRL suite~\citep{jaxgcrl} \{single-agent, continuous\}, SMAX~\citep{jaxmarl} and Connector~\citep{jumanji} \{multi-agent, discrete\} and  JaxNav~\citep{jaxnav} \{multi-agent, continuous\}. From these environment suites we test on a total of 18 tasks. The complete list of environments used and their descriptions are given in Appendix~\ref{app:environments}.

\paragraph{Baselines} 
We compare against baselines from the literature including several existing off-policy contrastive RL methods, here referred to as: CSAC~\citep{eysenbach2022} for single-agent continuous tasks, CDQN~\citep{demystifying-sgcrl} for single-agent discrete tasks and ICSAC~\citep{icrl} for multi-agent tasks. We note an important change. As mentioned in Table~\ref{tab:method-comparison}, our naming differs from the original names given to these approaches. For example, \citet{eysenbach2022} refer to their approach as CRL, while follow-up work by \citet{demystifying-sgcrl} propose a
discrete version of the same approach but instead of following a SAC-style optimisation, they closely follow a DQN-style algorithm without providing an explicit
name for it. Our aim with the above naming convention is to remove ambiguity in the naming by connecting the underlying base RL algorithm to the contrastive learning approach. 

\paragraph{Evaluation protocol and hyperparameters}
Unless otherwise specified, each algorithm is trained for 10 independent trials per task with a fixed budget of 80 million environment steps. We evaluated at 80 evenly spaced intervals with 2048 episodes per evaluation, recording the mean win rate in line with the recommendations from ~\cite{gorsane-protocol}. For per-task results, we report the mean with 95\% confidence intervals. For environment level aggregations, we report the min--max normalised inter-quartile mean (IQM) following ~\cite{rliable}. All per-task results can be found in Appendix~\ref{app:per-task-eval}.
Our evaluation aggregations, metric calculations, and plotting leverage the MARL-eval library ~\citep{gorsane-protocol}.

We obtained our hyperparameters from prior work when available ~\citep{jaxgcrl,sable}. When these were not available or the parameters provided performed poorly, we obtained them through a hyperparameter sweep using the Tree-structured Parzen Estimator (TPE) Bayesian optimization algorithm from the Optuna library~\citep{optuna}. For details on hyperparameters, we refer the reader to Appendix \ref{app:hyperparams}.

\subsection{Empirical Results and Discussion}

\newcommand{\panelW}{0.26\textwidth}     
\newcommand{\panelGap}{0pt}              
\newcommand{\rowGap}{0pt}                 
\newcommand{\hdrGap}{0pt}                 

\begin{figure}
    \centering
    \includegraphics[width=0.75\textwidth]{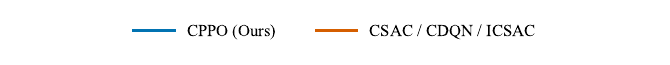}\par\vspace{0pt}

    \begin{minipage}{\textwidth}
        \centering
        {\small Single Agent}\par\vspace{\hdrGap}
        \includegraphics[width=\panelW]{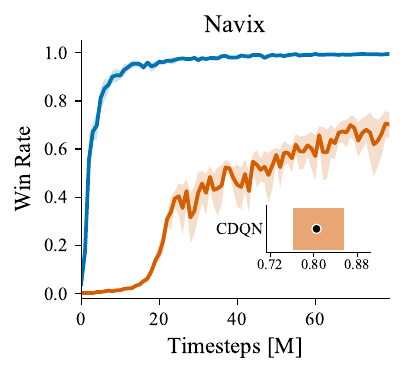}\hspace{\panelGap}%
        \includegraphics[width=\panelW]{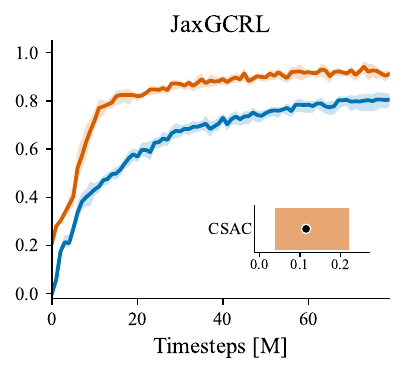}
    \end{minipage}\par\vspace{\rowGap}

    \begin{minipage}{\textwidth}
        \centering
        {\small Multi Agent}\par\vspace{\hdrGap}
        \includegraphics[width=\panelW]{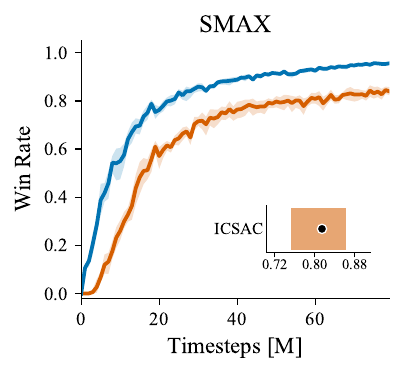}\hspace{\panelGap}%
        \includegraphics[width=\panelW]{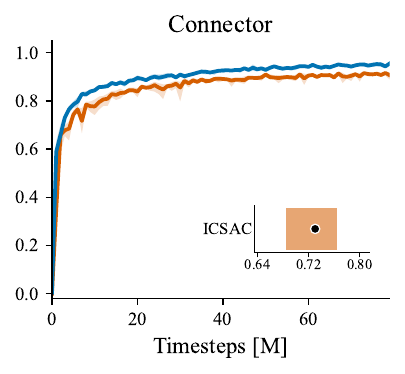}\hspace{\panelGap}%
        \includegraphics[width=\panelW]{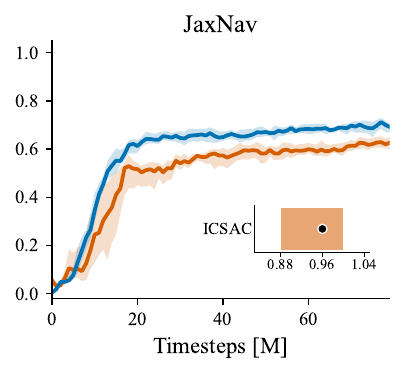}
    \end{minipage}

    \caption{\textbf{\textit{Per-environment IQM sample-efficiency curves}} (shaded 95\% CI) with inset probability-of-improvement bars, i.e. $P(\text{CPPO} > \text{baseline})$. Results are aggregated over multiple tasks from each environment suite. CPPO achieves higher mean performance than CRL baselines in 4/5 environments.}
    \label{fig:combined-grid}
\end{figure}


We organise our empirical investigation and discussion around four key questions: 
\begin{enumerate}
    \item How does our method compare to existing off-policy CRL?
    \item Can CPPO compete with hand-crafted dense rewards?
    \item How sensitive is reward design compared to goal specification?
    \item Does CPPO's performance scale with environment complexity? 
\end{enumerate}

\paragraph{How does CPPO compare to existing off-policy CRL?}


Figure~\ref{fig:combined-grid} and Table~\ref{tab:aggregate} show per-environment IQM sample-efficiency curves and win rates comparing CPPO/ICPPO against existing off-policy contrastive methods. Figure~\ref{fig:combined-grid} insets also show the probability of improvement of CPPO/ICPPO compared to the baseline. Per-task breakdowns are provided in Appendix~\ref{app:per-task-eval}. 

When comparing the off-policy and on-policy approaches, we find that
CPPO significantly outperforms the existing CRL baselines in all settings except the single-agent continuous control benchmark. CPPO natively supports discrete actions, without requiring a Gumbel-Softmax approximation as in \citet{icrl} or reverting to DQN as in \citet{demystifying-sgcrl}. Gumbel-Softmax introduces a biased continuous relaxation of the discrete action distribution, while reverting to DQN forgoes a parameterised policy and the actor-critic structure that most of CRL has been built around. We hypothesise both workarounds weaken the resulting CRL baseline relative to CPPO's more natural fit for the discrete setting.


We attribute CPPO's worse performance in single-agent continuous environments to two factors: First, in the discrete case, CPPO computes the state value exactly as an expectation over all actions (Equation~\ref{eq:value-discrete}); in the continuous case, this is intractable and must be approximated by Monte-Carlo sampling from the policy (Equation~\ref{eq:value-continuous}). We suspect this approximation carries higher variance than the $Q$ estimates used specifically by CSAC~\citep{eysenbach2022}. Second, CSAC builds on SAC, which consistently outperforms PPO on continuous control benchmarks~\citep{sac,sac-applications,open-rl-bench}, and has been refined through a long line of work in this setting~\citep{eysenbach2022,stabilizing-crl,jaxgcrl,1000layer}. 

A question arises from this, if CPPO struggles in continuous control, why does it outperform ICSAC in JaxNav, a continuous action space, multi-agent environment? We hypothesise that using the PPO backbone, which is considered state-of-the-art in MARL~\citep{sable}, helps CPPO perform well in these tasks. 
In summary, CPPO is the stronger contrastive RL method in the discrete and multi-agent settings, while off-policy contrastive RL retains an advantage in single-agent continuous control.

\begin{table}
\centering
\caption{\textbf{\textit{Per-environment aggregate IQM scores}} (normalised win rate / success rate) with 95\% stratified bootstrap confidence intervals in brackets. }
\label{tab:aggregate}
\resizebox{0.9\textwidth}{!}{%
\begin{tabular}{llll |cc}
\toprule
 & & & & \textbf{CPPO (Ours)} & \textbf{CSAC/CDQN/ICSAC} \\
\midrule
Navix & \{\texttt{4 tasks}\} & \texttt{discrete} & \texttt{single-agent}      & \textbf{0.995} {\scriptsize [0.993, 0.996]} & 0.702 {\scriptsize [0.646, 0.749]} \\
JaxGCRL & \{\texttt{3 tasks}\} & \texttt{continuous} & \texttt{single-agent} & 0.806 {\scriptsize [0.774, 0.832]} & \textbf{0.916} {\scriptsize [0.902, 0.930]} \\
SMAX & \{\texttt{6 tasks}\} & \texttt{discrete} & \texttt{multi-agent}      & \textbf{0.956} {\scriptsize [0.951, 0.960]} & 0.840 {\scriptsize [0.824, 0.859]} \\
Connector & \{\texttt{4 tasks}\} & \texttt{discrete} & \texttt{multi-agent}  & \textbf{0.955} {\scriptsize [0.950, 0.961]} & 0.909 {\scriptsize [0.899, 0.918]} \\
JaxNav &  \{\texttt{1 task}\} & \texttt{continuous} & \texttt{multi-agent}   & \textbf{0.692} {\scriptsize [0.672, 0.705]} & 0.626 {\scriptsize [0.600, 0.653]} \\
\bottomrule

\end{tabular}%
}
\end{table}
\paragraph{Can CPPO compete with hand-crafted dense rewards?}
\begin{wrapfigure}{r}{0.45\columnwidth}
    \centering
    \vspace{-10pt}
    \includegraphics[width=0.45\columnwidth]{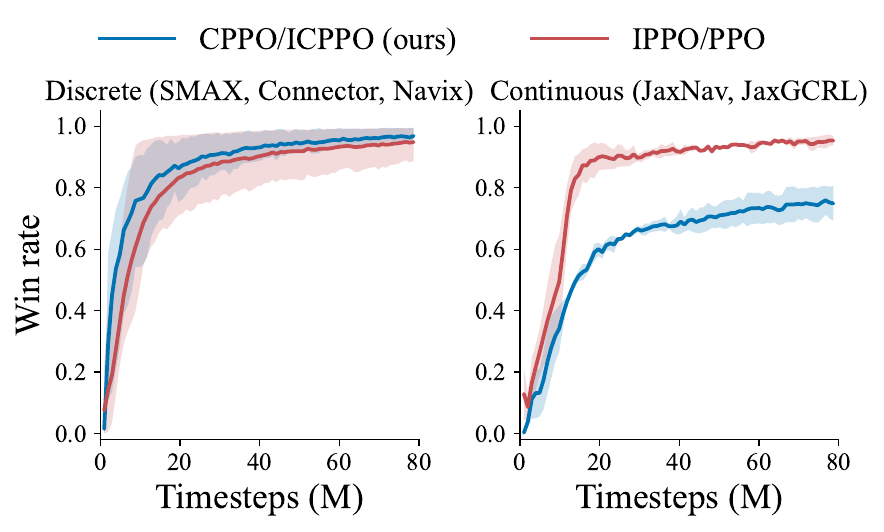}
    \captionsetup{font=small}
    \caption{\textbf{\textit{CPPO vs PPO with hand-crafted dense rewards}}, aggregated across discrete and continuous domains. In discrete settings CPPO matches or exceeds IPPO/PPO; in continuous settings a gap remains.}
    \label{fig:dense-comparison}
    \vspace{-9pt}
\end{wrapfigure}

Prior contrastive RL methods inherit the goal-conditioned formulation in which the reward is sparse by construction (Equation~\ref{eq:gcrl-objective}). Consequently, most of the work in CRL benchmarks performance almost exclusively against sparse-reward baselines~\citep{eysenbach2022,c-learning,stabilizing-crl,jaxgcrl,single-goal-crl}.
This is fair given the problem formulation, but in many real-world settings, a sensible reward function might be available, or could be designed with effort. In such cases, a practitioner would want the method that performs the best, regardless of whether it is self-supervised or uses a dense reward signal. We therefore evaluate CPPO against the hand-crafted dense rewards specifically designed for each environment in our study as originally proposed. Figure~\ref{fig:dense-comparison} shows the aggregate comparison across discrete and continuous environments, respectively. In discrete environments, CPPO matches I/PPO on SMAX and Navix and exceeds it on Connector, all without any reward signal. The gap reverses in continuous environments (JaxNav and JaxGCRL). This provides further evidence that the variance in Monte Carlo value estimation for CPPO in continuous settings might be negatively affecting performance.

This result is practically significant. It suggests that for discrete domains, practitioners could potentially bypass reward engineering entirely, a process that is brittle and environment-specific, with some confidence that they will not be sacrificing performance.

\paragraph{How sensitive is reward design compared to goal design?}
\label{sec:reward-ablation}

\newlength{\ablW}\setlength{\ablW}{0.235\textwidth}  
\newlength{\ablGap}\setlength{\ablGap}{2pt}            
\newlength{\ablHdrGap}\setlength{\ablHdrGap}{1.5pt}    
\begin{figure}
    \centering
    \begin{minipage}{\textwidth}
        \centering
        \makebox[\dimexpr 2\ablW + \ablGap\relax][c]{\small Connector}%
        \hspace{2\ablGap}%
        \makebox[\dimexpr 2\ablW + \ablGap\relax][c]{\small SMAX}\par
        \vspace{\ablHdrGap}
        \includegraphics[width=\ablW]{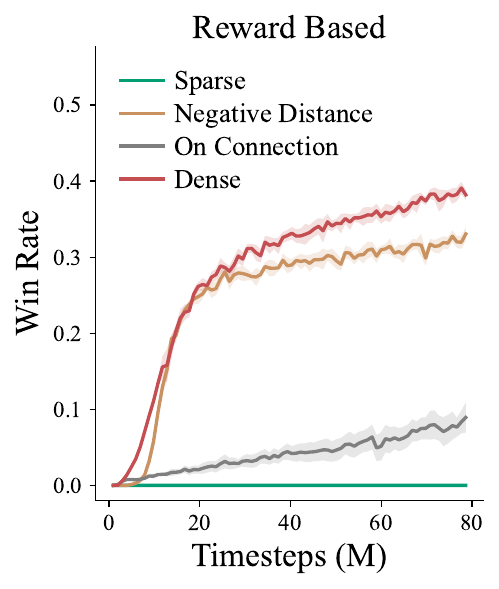}\hspace{\ablGap}%
        \includegraphics[width=\ablW]{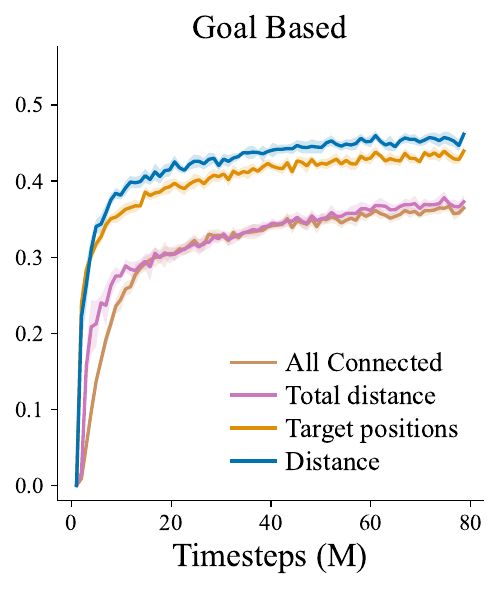}\hspace{2\ablGap}%
        \includegraphics[width=\ablW]{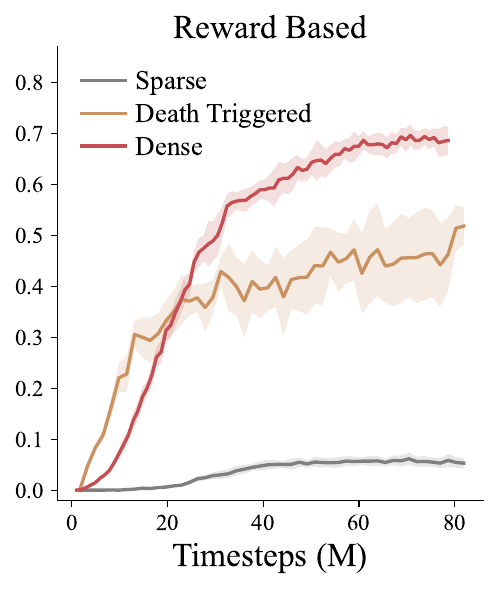}\hspace{\ablGap}%
        \includegraphics[width=\ablW]{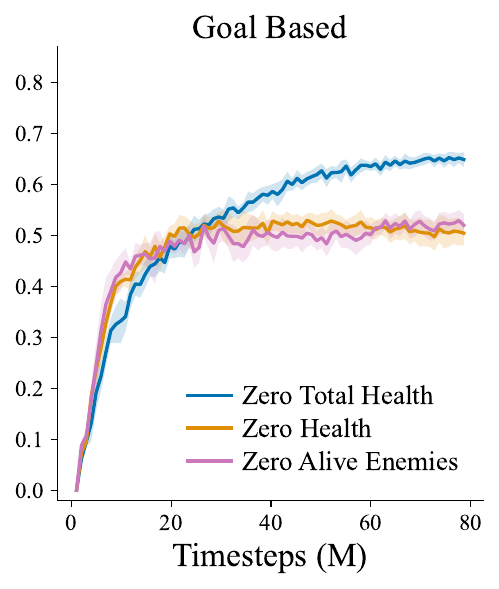}
    \end{minipage}

    \caption{Reward- vs Goal-design sensitivity on Connector 10×10 and SMAX (smacv2\_10\_units). For each environment, the reward panel shows IPPO trained under several hand-crafted reward variants, and the goal panel shows ICPPO trained under different goal representations. The reward range far exceeds the goal spread on both environments.}
    \label{fig:reward-goal-ablation}
\end{figure}

In practice, both reward-based and contrastive methods require design. Useful reward functions often require careful shaping, and contrastive methods require a sensible goal state definition. To quantify the sensitivity of these methods to misspecification, we select the Connector $10{\times}10$ and Smax V2 10 agent tasks and train IPPO under seven different, but all arguably sensible, reward functions, ranging from dense per step rewards to sparse rewards only provided on success. We run the same analysis for goal design and train ICPPO with seven goals of varying granularity from target positions to full connectivity. Full descriptions of each reward and goal are given in Appendix~\ref{app:reward-goal-desc}. 

Figure~\ref{fig:reward-goal-ablation} shows that the range from the best to worst reward is far greater than the range from the best to worst goal. It is important to note that we specifically included goals far more coarse than the default, which still perform relatively well. This means that goal specification is far more forgiving than reward design: even a coarse goal representation works reasonably well, whereas small changes to the reward function can cause large drops in performance.

\paragraph{Does CPPO's performance scale with environment complexity?}

\begin{wrapfigure}{r}{0.5\columnwidth}
    \centering
    \vspace{-10pt}
    \includegraphics[width=0.48\columnwidth]{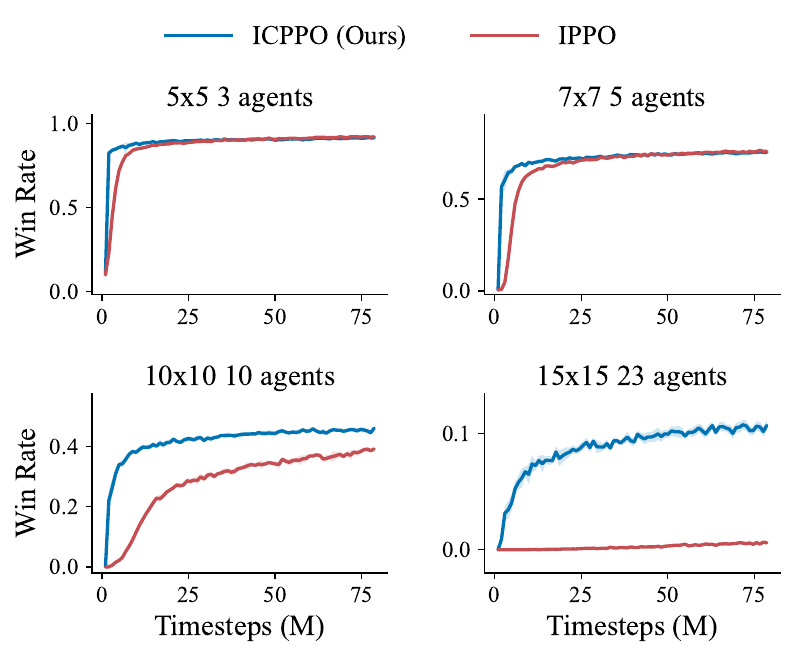}
    \captionsetup{font=small}
    \caption{Connector scaling: per-task learning curves across four grid sizes. All methods are similar at $5{\times}5$; ICPPO's advantage widens as coordination complexity grows.}
    \label{fig:connector-scaling}
    \vspace{-10pt}
\end{wrapfigure}
We compare CPPO and IPPO on four Connector variants of increasing size and agent count: $5{\times}5$ (3 agents), $7{\times}7$ (5 agents), $10{\times}10$ (10 agents), and $15{\times}15$ (23 agents). Figure~\ref{fig:connector-scaling} shows the per-task learning curves.
At the smallest scale ($5{\times}5$) both methods converge to similar win rates (${\sim}91\%$). As the grid grows, reward-based methods degrade faster: at $10{\times}10$, ICPPO reaches 46.2\% vs.\ IPPO's 39.3\%. At $15{\times}15$ the gap is most pronounced as ICPPO attains 10.7\% while IPPO collapses to 0.6\%. 

Two properties of the environment plausibly drive this divergence. First, optimal trajectories lengthen with grid size, so bootstrapped value targets must propagate across more steps thus accumulating more error; a contrastive objective supervises returns directly rather than chaining through bootstrapped estimates, sidestepping this compounding. Second, successful connections become rarer as the task grows harder, so the signal a critic regresses against when following a mean squared error objective becomes increasingly sparse and dominated by near-zero returns; a contrastive critic only needs to distinguish better trajectories from worse ones and remains informative even when absolute reward magnitudes carry little signal. These properties are not unique to Connector and suggest that the contrastive critic provides a more useful learning signal in complex environments, even when MSE-based value regression suffices at small scale.

\section{Related Work}

\paragraph{Contrastive reinforcement learning.}
Goal-conditioned RL has historically relied on hindsight relabeling and incentivising goal-reaching behaviour via sparse rewards~\citep{Kaelbling,her,hindsight-inverse-dynamics,imagined-subgoals,shes}. \citet{eysenbach2022} showed that contrastive learning on action-labled trajectories yields goal-conditioned Q-functions, unifying CRL and GCRL. Recent work has addressed practical challenges by introducing JAX-based implementations of GCRL benchmarking environments and algorithms~\citep{jaxgcrl}, scaled to large networks~\citep{1000layer}, introduced offline variants~\citep{stabilizing-crl,og-bench}, added TD bootstrapping~\citep{td-infonce}, demonstrated emergent exploration~\citep{demystifying-sgcrl} and extended algorithms to the coooperative multi-agent case~\citep{icrl}. All contrastive methods rely on off-policy optimisation. 

\paragraph{On-policy policy gradient methods.}
PPO~\citep{ppo}, a practical trust region method and clipped-surrogate successor to TRPO~\citep{trpo}, is a widely used on-policy algorithm in modern RL. It improves over actor-critic methods like A2C~\citep{a2c} and uses generalised advantage estimation~\citep{gae} to balance the bias variance trade-off. It has become even more popular due to massively parallel simulators~\citep{isaacgym,gymnax,jumanji} where high throughput outweighs the sample efficiency of off-policy methods. Furthermore trust region methods remain a target for theoretical~\citep{mirror-learning} and algorithmic~\citep{dpo} refinement. Yet existing contrastive RL methods are off-policy~\citep{eysenbach2022,demystifying-sgcrl,1000layer}. CPPO closes this gap.

\paragraph{Cooperative multi-agent RL.}

Cooperative MARL is broadly split into two paradigms: independent learning, where each agent independently optimises its own policy from local observations~\citep{ippo}, and centralised training with decentralised execution~\citep{ctde}, which exploits global information at training time~\citep{maddpg,vdn,qmix,mappo}. Heterogeneous-agent variants~\citep{happo} introduce monotonic improvement guarantees and are extended by sequence-modeling approaches~\citep{MAT, sable}. All rely on hand-designed dense team rewards. ICRL~\citep{icrl}, the sole multi-agent contrastive extension builds on SAC, which is at odds with cooperative MARL, where on-policy methods are typically preferred~\citep{mappo}. ICPPO is its on-policy counterpart.
\section{Conclusion}
\label{sec:conclusion}

We introduced CPPO, an on-policy contrastive reinforcement learning algorithm that derives advantages directly from contrastive Q-values and optimises them with PPO's clipped surrogate objective. The result is a single self-supervised algorithmic template that operates across discrete and continuous action spaces and across single- and multi-agent settings, integrating naturally into the on-policy pipelines widely used in modern industrial RL applications.

Empirically, CPPO matches or exceeds reward-engineered PPO and IPPO on every discrete benchmark we evaluate, despite using no reward signal, and surpasses prior contrastive baselines in every setting except single-agent continuous control. Our analysis further illustrates how goal design can be markedly more forgiving than reward design and that CPPO's advantage over reward-based baselines widens as environment complexity grows.

\paragraph{Limitations and future work} A clear limitation remains that in single-agent continuous control, CPPO trails off-policy SAC-based contrastive methods. We hypothesise that this gap may stem from variance introduced by the Monte Carlo state-value estimate, which off-policy SAC-based contrastive methods avoid. However, we do not provide any evidence to substantially support this claim. Further analysis that could point to potential algorithmic improvements for the continuous setting is a natural direction for future work. While the multi-agent ICPPO performs remarkably well, a centralised training with decentralised execution version is another promising avenue for further investigation.

\newpage
\bibliographystyle{plainnat}
\bibliography{refs}

\appendix  
\newpage
\section{Environment Details}
\label{app:environments}

We provide a short conceptual description of each environment.  Implementation
details (exact observation layouts, hyperparameters, code-level flags) are
available in the original references.  Reward and goal variants used in our
ablations or not mentioned here are described separately in
Appendix~\ref{app:reward-goal-desc}.

\subsection{SMAX}
\label{app:env:smax}

SMAX~\citep{jaxmarl} is a JAX-reimplementation of the StarCraft Multi-Agent
Challenge, with SMACv2 scenarios that randomise unit composition and
starting positions~\citep{smacv2}.  Teams of allied units must coordinate
to defeat an enemy team controlled by a hand-coded heuristic that attacks
the nearest visible opponent.

\paragraph{Observation space.}
Each agent receives a partial observation limited by a unit-type-dependent
sight range, encoding its own state and the relative state (position,
health, unit type) of allied and enemy units within range.  A binary mask
indicates which actions are currently legal.

\paragraph{Action space.}
Discrete: four cardinal movement directions, stop, no-op, and one
``attack enemy~$i$'' action per enemy unit.

\paragraph{Scenarios.}
We evaluate on six scenarios: \textbf{3m} and \textbf{8m} (symmetric
homogeneous matchups), \textbf{5m\_vs\_6m} (asymmetric, requires
focus-fire), \textbf{6h\_vs\_8z} (heterogeneous, requires kiting), and
the SMACv2 scenarios \textbf{SMACv2 5\,units} and \textbf{SMACv2
10\,units}, which randomise unit types and starting positions each
episode.
\paragraph{Reward.}
The default dense reward equally incentivises tactical engagement and
overall victory: agents earn $50\%$ of their return from per-step damage
events (damage dealt minus damage received) and $50\%$ from winning the
episode. 

\paragraph{Goal definition.}
The goal is the elimination of the enemy team. Following~\citet{icrl},
ICPPO conditions on a goal corresponding to this end-state, a scalar representing
the (normalised) total enemy health, with target value zero (all enemies
eliminated).

\subsection{Connector}
\label{app:env:connector}

The Vector Connector environment~\citep{jumanji} is a
cooperative multi-agent grid world in which each agent must trace a
connected path from its designated start cell to its target cell without
overlapping the paths of other agents.

\paragraph{Observation space.}
Each agent observes its own position and target, an egocentric local view
of nearby agents and their trail cells, and an egocentric view of all
agents' targets.  

\paragraph{Action space.}
Five discrete actions: move in one of the four cardinal directions, or
stay in place.

\paragraph{Tasks.}
We evaluate on four configurations: $5{\times}5$ grid with 3~agents,
$7{\times}7$ with 5~agents, $10{\times}10$ with 10~agents, and
$15{\times}15$ with 23~agents.  The episode horizon scales with grid size
(set to the number of grid cells, capped at $T{=}225$ for the largest).

\paragraph{Reward.}
The default reward grants $+1$ when an agent connects its
endpoints and applies a $-0.03$ per-step penalty otherwise.  An episode
is ``won'' when every agent has connected.  

\paragraph{Goal definition.}
The goal is for each agent to reach its assigned target cell, completing
its individual connection. ICPPO/ICRL uses a default of per-agent normalised Manhattan distance from current to target position, with target value zero (agent at its endpoint). 

\subsection{JaxNav}
\label{app:env:jaxnav}

JaxNav~\citep{jaxnav,jaxmarl} is a JAX-native 2D continuous-space
navigation environment for differential-drive robots.  Each robot must
reach an individual goal position on a randomly generated cluttered map
while avoiding walls and the other robots.

\paragraph{Observation space.}
Continuous.  Each robot observes a vector of LiDAR range readings sampled
over a $360^\circ$ arc, the polar coordinates of its goal relative to its
current pose, and its own linear and angular velocity.

\paragraph{Action space.}
Continuous and two-dimensional: a target linear velocity and target
angular velocity, integrated through differential-drive kinematics.

\paragraph{Reward.}
The default reward combines a goal-arrival bonus, a distance-shaping
term that rewards moving closer to the goal, a collision penalty (walls
or other robots), a proximity penalty for near-collisions, and a small
per-step time penalty.  See \citet{jaxnav} for the full equations.

\paragraph{Goal definition.}
The goal is for the robot to arrive at its assigned target location
without colliding with obstacles or other robots. Our contrastive experiments use the Euclidean distance from the agent's current position to its goal, with a target value of zero (agent at its goal).

\subsection{JaxGCRL}
\label{app:env:jaxgcrl}

The JaxGCRL suite~\citep{jaxgcrl} is a JAX-native benchmark of
single-agent continuous-control goal-reaching tasks built on the Brax
physics engine.  We evaluate on three tasks: \emph{Reacher}, \emph{Ant},
and \emph{Ant U-Maze}.  We refer the reader to \citet{jaxgcrl} for full
implementation details and hyperparameter defaults.

\paragraph{Reacher.}
A planar two-link arm whose end-effector must reach a randomly sampled
target position.  Observations are continuous and include the joint
angles, end-effector position, and end-effector linear velocity; actions
are continuous joint torques.  A run is successful when the
end-effector is within a small radius of the target.  The default reward
is the negative Euclidean distance from end-effector to target. 

\paragraph{Ant.}
A quadruped (eight hinge actuators) that must reach a randomly sampled
goal position on a flat plane.  Observations include the torso pose,
joint angles, and their velocities; actions are continuous joint
torques.  Goals are sampled at a fixed distance from the spawn with a
uniformly random angle, and a run succeeds when the torso is within
$0.5$~units of the goal.  The default reward combines a
velocity-toward-goal term, a healthy-pose bonus, and a control penalty.

\paragraph{Ant U-Maze.}
The same Ant morphology placed inside a U-shaped corridor on a
$5{\times}5$ cell grid.  The agent spawns at a fixed reset cell and must
reach a goal cell sampled uniformly from the six valid goal cells.  The
shortest path requires navigating \emph{around} the central wall, making
straight-line heuristics misleading.  Observation, action, success
criterion, and reward formulation are identical to the flat-ground Ant
task.
\paragraph{Goal definition.}
The goal in each task is the target end-state: in Reacher, the
end-effector positioned at the sampled target; in Ant, the torso
positioned at the sampled goal location; in Ant U-Maze, the torso at
one of possible goal cells inside the U-shaped maze, sampled
uniformly each episode. CPPO/CSAC conditions on these end-states, a
slice of the observation corresponding to the relevant body part
position (the end-effector for Reacher, the root $(x,y)$ for Ant and
U-Maze), with target value equal to the sampled goal.


\subsection{Navix}
\label{app:env:navix}

Navix~\citep{navix} is a JAX-native reimplementation of the MiniGrid
grid-world suite.  An agent occupies a single cell on a discrete grid
and must navigate to a designated goal cell, optionally completing
intermediate sub-goals such as picking up a key and toggling a door.

\paragraph{Observation space.}
Under partial observability the agent receives a $7{\times}7{\times}3$
egocentric crop aligned with its facing direction, where the three
channels follow MiniGrid's symbolic encoding (object tag, colour, and
state) per cell.

\paragraph{Action space.}
Seven discrete actions following the MiniGrid default set: rotate
counter-clockwise, rotate clockwise, move forward, pick up, drop,
toggle, and \texttt{done}.

\paragraph{Tasks.}
We evaluate on four scenarios:
\textbf{Empty-16$\times$16} (empty room, fixed start/goal),
\textbf{Empty-Random-16$\times$16} (empty room, randomised start and
goal each episode),
\textbf{DoorKey-Random-16$\times$16} (locked door with randomised
key/door/start/goal positions; the agent must pick up the key, toggle
the door open, and reach the goal), and
\textbf{FourRooms} (four-room layout with narrow doorways; the agent
must traverse multiple rooms to reach a randomised goal).

\paragraph{Reward.}
Navix's default reward is sparse: $+1$ on reaching the goal, $0$
otherwise. Our PPO baseline uses a lightly-shaped composite reward
built from Navix primitives: $+1$ on reaching the goal, $-0.01$ per
environment step, and $-0.01$ for each wall collision. The shaping
terms are two orders of magnitude smaller than the goal bonus, so the
signal remains close to sparse; in our experiments, PPO achieves the
same final performance under the purely sparse variant.
\paragraph{Goal definition.}
The goal is for the agent to reach its assigned goal cell. On Empty
and FourRooms, we encode it as the goal cell's $(y, x)$ grid
coordinates (2D); the contrastive Q-function learns to encode integer
grid positions directly. On DoorKey the agent must first pick up a
key and toggle a door open before reaching the goal; we encode it as
a 1D scalar progress signal in $[0, 1]$ that combines the key-to-door
and player-to-goal Manhattan distances, with target value one (full
progress).

\section{Goal and Reward Designs}
\label{app:reward-goal-desc}

This appendix lists the goal and reward variants used in our
goal- and reward-design ablations (Section~\ref{sec:reward-ablation}).
We tested variants on Connector and SMAX; the default goals and
rewards for the other environments are described in
Appendix~\ref{app:environments}.

\subsection{Connector}
\label{app:reward-goal-desc:connector}

\paragraph{Reward variants.}
We compare four reward functions on Connector spanning a range of
informativeness. \emph{Dense} and \emph{Negative Distance} provide a
per-step gradient that pushes each agent toward its target.
\emph{On Connection} fires a one-shot $+1$ on the step an agent connects.
\emph{Sparse} is the strictest: it fires only when every
agent is simultaneously connected.

\begin{table}[ht!]
\centering
\caption{Connector reward variants used in the reward sensitivity ablation.
Per-agent rewards are summed across agents and broadcast as the team total, as is the default in Mava.}
\label{tab:reward-variants-connector}
\begin{tabular}{ll}
\toprule
\textbf{Name} & \textbf{Per-agent reward} \\
\midrule
Dense             & \begin{tabular}[t]{@{}l@{}}$+1$ on the step the agent connects;\\$-0.03$ on every step it remains unconnected.\end{tabular} \\[6pt]
Negative Distance & \begin{tabular}[t]{@{}l@{}}Negative Manhattan distance from the agent's\\position to its target.\end{tabular} \\[6pt]
On Connection            & \begin{tabular}[t]{@{}l@{}}$+1$ on the single step the agent connects;\\zero otherwise.\end{tabular} \\[6pt]
Sparse   & \begin{tabular}[t]{@{}l@{}}$+1$ to every agent only when all agents are\\simultaneously connected; zero at all other steps.\end{tabular} \\
\bottomrule
\end{tabular}
\end{table}

\paragraph{Goal variants.}
We compare four goal representations on Connector spanning a range
of granularity. \emph{Distance} (default) gives each agent its own
progress signal toward its individual target. \emph{Total distance}
collapses this to a single team-mean signal broadcast to every agent.
\emph{All Connected} reports only how many agents have reached their
endpoints. \emph{Target positions} uses the target cell's $(x, y)$
coordinates as the goal vector, with the agent's own position as the
achieved state.

\begin{table}[ht!]
\centering
\caption{Connector goal variants used in the goal sensitivity ablation.}
\label{tab:goal-variants-connector}
\begin{tabular}{ll}
\toprule
\textbf{Name} & \textbf{Description} \\
\midrule
Distance (default) & \begin{tabular}[t]{@{}l@{}}Per-agent normalised Manhattan distance to target;\\target value zero (agent at its endpoint).\end{tabular} \\[6pt]
Total distance     & \begin{tabular}[t]{@{}l@{}}Team-mean of the per-agent distances, broadcast\\to every agent; target value zero.\end{tabular} \\[6pt]
All Connected      & \begin{tabular}[t]{@{}l@{}}Fraction of agents currently connected, broadcast\\to every agent; target value one.\end{tabular} \\[6pt]
Target positions   & \begin{tabular}[t]{@{}l@{}}Per-agent: $(x, y)$ position as achieved state,\\target $(x, y)$ as goal vector.\end{tabular} \\
\bottomrule
\end{tabular}
\end{table}

\subsection{SMAX}
\label{app:reward-goal-desc:smax}

\paragraph{Reward variants.}
We compare three reward functions on SMAX spanning a range of
informativeness. All three share the same terminal win/loss bonus;
they differ only in the per-step shaping signal. SMAX's default
\emph{Dense} reward~\citep{jaxmarl} provides per-step shaping
proportional to the normalised damage dealt to enemies. Our
\emph{Death Triggered} variant replaces this continuous damage signal
with a discrete bonus that fires only when an enemy is killed. This
is a sub-component analogous to the \texttt{reward\_death\_value}
term in the original SMAC~\citep{smacv2}. \emph{Sparse} drops both shaping
terms entirely, leaving only the terminal win/loss bonus.

\begin{table}[ht!]
\centering
\caption{SMAX reward variants used in the reward sensitivity ablation.
All variants additionally include a terminal win/loss bonus.}
\label{tab:reward-variants-smax}
\begin{tabular}{ll}
\toprule
\textbf{Name} & \textbf{Per-step shaping} \\
\midrule
Dense (default)  & \begin{tabular}[t]{@{}l@{}}Sum of normalised damage dealt to enemies\\at each step, $+1$ if episode won.\end{tabular} \\[6pt]
Death Triggered  & \begin{tabular}[t]{@{}l@{}}Bonus of $R_{\text{win}}/num\_enemies$ on each step an enemy is\\killed, $+1$ if episode won;\\zero per-step otherwise.\end{tabular} \\[6pt]
Sparse           & \begin{tabular}[t]{@{}l@{}}No per-step shaping; reward only at the\\terminal win/loss event.\end{tabular} \\
\bottomrule
\end{tabular}
\end{table}

\paragraph{Goal variants.}
We compare three goal representations on SMAX, all formalising the
same end-state (defeat the enemy team) at different granularities. \emph{Zero Total Health} (default) follows~\citet{icrl}
and uses a single scalar that aggregates the normalised health of all
enemies. \emph{Zero Health} replaces this with a per-enemy vector, one
entry per enemy unit. \emph{Zero Alive Enemies} collapses the signal
to a single team-level fraction over currently-observable enemies.
All three rely on per-agent visibility: enemies that are not currently
observed are assumed to be at full health (i.e., still alive).

\begin{table}[ht!]
\centering
\caption{SMAX goal variants used in the goal sensitivity ablation.}
\label{tab:goal-variants-smax}
\begin{tabular}{ll}
\toprule
\textbf{Name} & \textbf{Description} \\
\midrule
Zero Total Health (default) & \begin{tabular}[t]{@{}l@{}}Scalar sum of normalised enemy healths;\\target value zero (all enemies eliminated).\end{tabular} \\[6pt]
Zero Health                 & \begin{tabular}[t]{@{}l@{}}Per-enemy vector of normalised damage values\\(one entry per enemy); target corresponds to\\all enemies eliminated.\end{tabular} \\[6pt]
Zero Alive Enemies          & \begin{tabular}[t]{@{}l@{}}Scalar fraction of currently-visible enemies\\still alive; target value zero (no alive enemies\\in view).\end{tabular} \\
\bottomrule
\end{tabular}
\end{table}

\section{Benchmark}
\label{app:per-task-eval}
\begin{figure}
    \centering
    \includegraphics[width=0.95\textwidth]{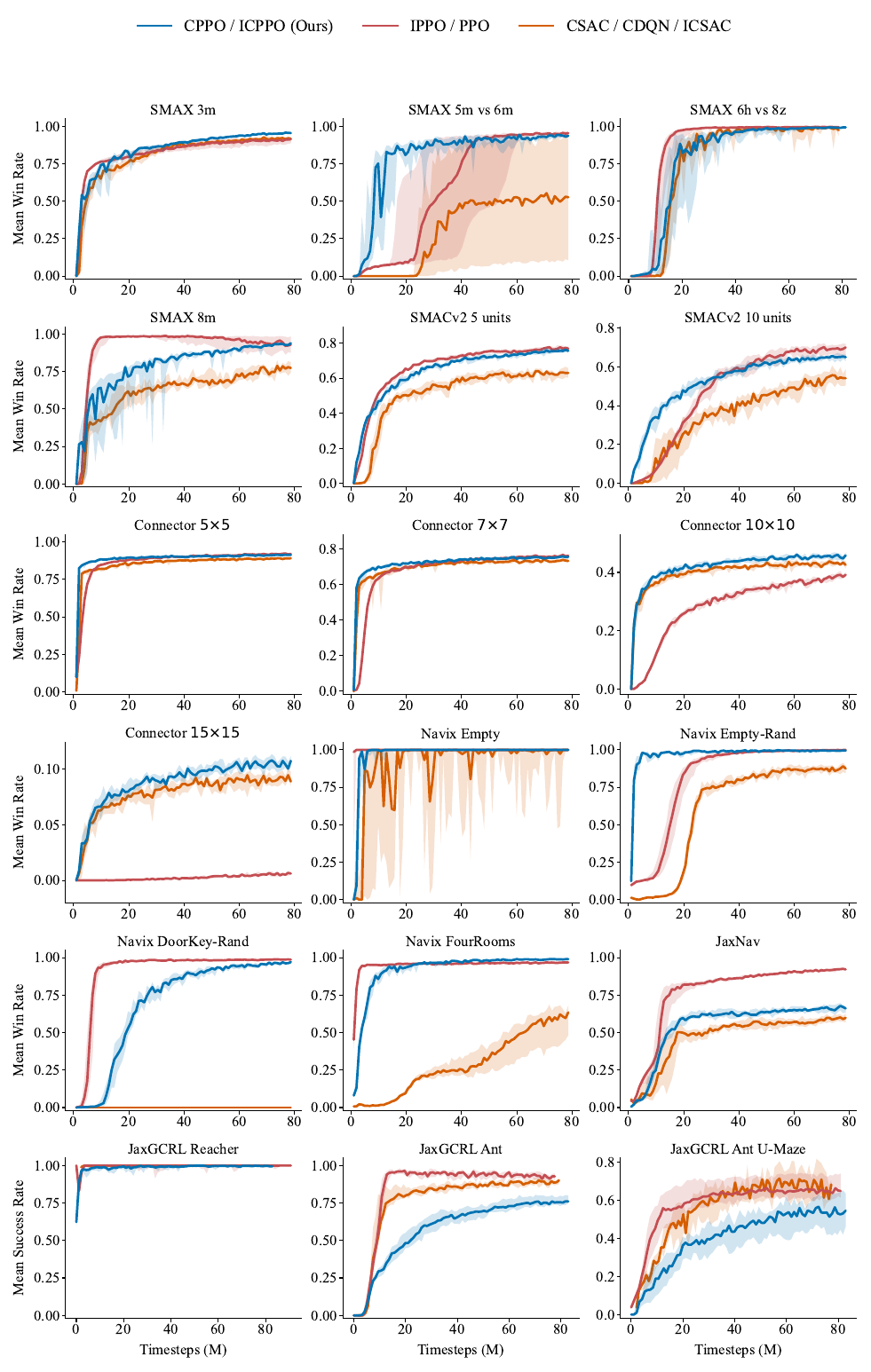}
    \caption{Mean Win Rate / Success Rate  with 95\% bootstrap confidence intervals on all tasks.}
    \label{fig:per-task-grid}
\end{figure}
The body of the paper reports per-environment IQM aggregates (Figure~\ref{fig:combined-grid} and Table~\ref{tab:aggregate}), here we present per-task learning curves below. Each panel shows the per-task IQM (mean of the middle 50\% of seeds at every evaluation step) over 10 seeds for the three method families: CPPO/ICPPO (ours, blue), the on-policy dense-reward baseline (PPO/IPPO, red), and the off-policy contrastive baseline of \citet{eysenbach2022,demystifying-sgcrl,icrl} (CSAC/CDQN/ICSAC, orange). Shaded bands are 95\% bootstrap confidence intervals over seeds.
Figure~\ref{fig:per-task-grid} provides a single-page grid of all 18 tasks for at-a-glance comparison.


\section{Hyperparameters}
\label{app:hyperparams}

This appendix documents the hyperparameters used in our experiments.
For our method (CPPO/ICPPO), we report the values that we held fixed
across all tasks. Table~\ref{tab:hp_cppo_global} lists the
architectural, contrastive, and training-scale defaults that apply
to every one of the $18$ tasks; Table~\ref{tab:hp_cppo_action_space}
lists the optimisation knobs whose recommended value depends on the
action space (discrete vs.\ continuous). For the baselines, we adopt
prior-work configurations when available and otherwise re-tune via
a per-task sweep, as detailed in Section~\ref{app:hp:baselines}. To ensure a fair comparison, the training-scale settings of Table~\ref{tab:hp_cppo_global} (rollout
length, batch size, parallel environments, total number of updates,
and evaluation budget) are matched across all algorithms, including
baselines. All experiments were run on NVIDIA H100 GPUs with 8 CPU cores
allocated per job. A single training seed of a single task took
between roughly 10 minutes (smallest tasks, e.g.\ SMAX 3m or
Connector $5{\times}5$) and 2 hours (largest tasks, e.g.\ Connector
$15{\times}15$ or SMACv2 10 units) to complete the full $80$M-step
training budget.
The remaining per-task tuned values
for our method, the full baseline configurations, and all sweep
specifications are released on the project
website\footnote{~\url{https://sites.google.com/view/contrastive-ppo/home}}.

\begin{table}[ht]
\centering
\caption{\textbf{Default hyperparameters for CPPO/ICPPO}}
\label{tab:hp_cppo_global}
\small
\setlength{\tabcolsep}{8pt}
\begin{tabular}{ll}
\toprule
\textbf{Hyperparameter} & \textbf{Value} \\
\midrule
\multicolumn{2}{l}{\emph{Network}} \\[2pt]
Hidden layer sizes                           & [512,\,512,\,512,\,512] \\
Activation                                   & swish \\
Layer normalisation                          & true \\
\midrule
\multicolumn{2}{l}{\emph{Contrastive}} \\[2pt]
Contrastive objective                        & forward InfoNCE \\
Energy function                              & L2 \\
Representation dimension                     & 64 \\
\midrule
\multicolumn{2}{l}{\emph{Training scale}} \\[2pt]
Rollout length                               & 128 \\
Batch size                                   & 256 \\
Number of updates                            & 1\,250 \\
Number of evaluations                        & 80 \\
Parallel environments                        & 512 \\
Evaluation episodes per evaluation           & 2\,048 \\
\bottomrule
\end{tabular}
\end{table}

\begin{table}[ht]
\centering
\caption{\textbf{Action-space-dependent default hyperparameters for
CPPO/ICPPO.} The discrete column covers SMAX (6), Connector (4)
and Navix (4); the continuous column covers JaxNav (1) and
JaxGCRL (3). Bracketed entries in the continuous column give the
$[\min,\max]$ range of selected values across the four tasks; the
exact per-task value is released on the project website.}
\label{tab:hp_cppo_action_space}
\small
\setlength{\tabcolsep}{8pt}
\begin{tabular}{lcc}
\toprule
\textbf{Hyperparameter} & \textbf{Discrete (14 tasks)} & \textbf{Continuous (4 tasks)} \\
\midrule
Gamma ($\gamma$)              & 0.99                 & \scriptsize{$[0.99,\,0.9999]$} \\
PPO clip ($\epsilon$)         & 0.2                  & \scriptsize{$[0.12,\,0.20]$} \\
PPO epochs                    & 1                    & \scriptsize{$\{1,\,8\}$} \\
Actor LR                      & $2.5{\times}10^{-4}$ & \scriptsize{$[1.7{\times}10^{-5},\,3{\times}10^{-4}]$} \\
$Q$ LR                        & $2.5{\times}10^{-4}$ & \scriptsize{$[3.9{\times}10^{-5},\,3{\times}10^{-4}]$} \\
Max gradient norm             & 0.5                  & \scriptsize{$[0.5,\,5]$} \\
LR decay schedule             & cosine               & \scriptsize{none,linear} \\
LR end (final value)          & $1{\times}10^{-7}$   & \scriptsize{$[1{\times}10^{-7},\,1{\times}10^{-6}]$} \\
\bottomrule
\end{tabular}
\end{table}

\subsection{Baseline hyperparameters}
\label{app:hp:baselines}

We adopt baseline hyperparameters from prior work whenever a
reference configuration is published for the corresponding task,
and re-tune via a per-task sweep otherwise, either because no
reference exists or because the published values failed to
reproduce the performance reported in the original work on our
benchmark. All sweeps use the Tree-structured Parzen Estimator
(TPE) Bayesian optimisation algorithm from
Optuna~\citep{optuna}.

We adopt published configurations directly for the following
baseline--environment pairs: IPPO on SMAX and Connector uses hyperparameters  published in ~\citet{sable}; IPPO on JaxNav uses the reference configuration released with the JaxNav
codebase~\citep{jaxnav}. PPO and CSAC on JaxGCRL inherit the canonical Brax recipes from ~\citet{jaxgcrl}; and ICSAC on SMAX uses the hyperparameters recommended by ~\citet{icrl} and reproduced results consistent with their released codebase

The remaining baseline-environment pairs are tuned via TPE
sweep. PPO and CDQN on Navix are tuned from scratch, and ICSAC on Connector and JaxNav is tuned on top of the~\citet{icrl} codebase using their published values as
the starting point.


\end{document}